\documentclass[runningheads]{llncs}
\usepackage{multirow}
\usepackage{amsmath}
\usepackage{amsfonts}
\usepackage{amssymb}
\usepackage{booktabs}
\usepackage{bbding}
\usepackage[T1]{fontenc}
\usepackage{hyperref}
\usepackage{graphicx,verbatim}
\begin{document}
\title{
Endo-CLIP: Progressive Self-Supervised Pre-training on Raw Colonoscopy Records}
%

\author{
Yili He\inst{1,2,3}\dag \and Yan Zhu\inst{4,5}\dag \and Peiyao u\inst{4,5} \and Ruijie Yang\inst{6} \and Tianyi Chen\inst{1,3} \and Zhihua Wang\inst{6} \and Quanlin Li\inst{4,5} \and Pinghong Zhou\inst{4,5} \and Xian Yang\inst{7,8}\Envelope \and Shuo Wang\inst{1,3,5,8}\Envelope}

\authorrunning{Y. He et al.}

\institute{
\textsuperscript{1}Digital Medical Research Center, School of Basic Medical Sciences, Fudan University, Shanghai, China \\
\inst{2}University College London, London, UK \\
\inst{3}Shanghai Key Laboratory of MICCAI, Shanghai, China \\
\inst{4}Endoscopy Center and Endoscopy Research Institute, Zhongshan Hospital, Fudan University, Shanghai, China \\
\inst{5}Shanghai Collaborative Innovation Center of Endoscopy, Shanghai, China \\
\inst{6}Shanghai Institute for Advanced Study of Zhejiang University, Shanghai, China \\
\inst{7}Alliance Manchester Business School, The University of Manchester, Manchester, UK \\
\inst{8}Data Science Institute, Imperial College London, London, UK
}

\maketitle              
\def\thefootnote{\dag}\footnotetext{Equal contribution.}
\def\thefootnote{\Envelope}\footnotetext{Corresponding authors: shuowang@fudan.edu.cn and xian.yang@manchester.ac.uk}

\begin{abstract}
Pre-training on image-text colonoscopy records offers substantial potential for improving endoscopic image analysis, but faces challenges including non-informative background images, complex medical terminology, and ambiguous multi-lesion descriptions.  We introduce Endo-CLIP, a novel self-supervised framework that enhances Contrastive Language-Image Pre-training (CLIP) for this domain. Endo-CLIP's three-stage framework—cleansing, attunement, and unification—addresses these challenges by: (1) removing background frames, (2) leveraging large language models (LLMs) to extract clinical attributes for fine-grained contrastive learning, and (3) employing patient-level cross-attention to resolve multi-polyp ambiguities. Extensive experiments demonstrate that Endo-CLIP significantly outperforms state-of-the-art pre-training methods in zero-shot and few-shot polyp detection and classification, paving the way for more accurate and clinically relevant endoscopic analysis. Code will be made publicly available on \url{https://github.com/chrlott/EndoCLIP}.


\end{abstract}

\section{Introduction}\label{sec:intro}
Colorectal cancer remains a leading cause of cancer-related deaths globally~\cite{arnold2017global}. Early detection and removal of precancerous polyps through colonoscopy significantly improves patient outcomes. Endoscopic examinations are crucial for this process, allowing clinicians to visualize the gastrointestinal tract and identify suspicious lesions. Computer-assisted systems have demonstrated significant improvements in adenoma detection rates (ADR)~\cite{le2020application}. These systems primarily target two key tasks~\cite{ali2022we}: polyp detection (identifying the presence of polyps) and optical biopsy (determining polyp histology in real-time during the procedure). While artificial intelligence (AI)-driven models have achieved near-expert performance, their generalization and robustness across diverse clinical settings and patient populations remain limited~\cite{ahmad2021establishing}, impeding adoption and clinical use.

Pre-training, a technique that learns generalizable visual representations from large datasets, has shown considerable promise in medical imaging \cite{zhang2023biomedclip,bao2021beit}. By pre-training image encoders on massive datasets, subsequent transfer learning to specific downstream tasks becomes more effective, even with limited labeled data \cite{dai2023swin}. The vast, daily-accumulating archives of electronic colonoscopy records – comprising numerous screenshots and free-text reports – represent a largely untapped resource for such pre-training. Self-supervised learning, particularly contrastive learning approaches like DINO \cite{caron2021emerging}, offers a powerful means to leverage this data without manual annotations \cite{he2020momentum,oquab2023dinov2,zhou2023foundation}. Contrastive Language-Image Pre-training (CLIP) \cite{radford2021learning} and its variants have demonstrated impressive success by aligning images and text in a shared embedding space. In the medical domain, adaptations like MedCLIP \cite{wang2022medclip} have successfully leveraged image-text pairings in clinical settings, notably in X-ray diagnosis~\cite{zhang2022contrastive,tiu2022expert,lin2023pmc,chen2023knowledge}, ultrasound~\cite{christensen2024vision} and digital pathology~\cite{xu2024whole,huang2023visual}.

\begin{figure}[h!]
\centering
\includegraphics[scale=0.38]{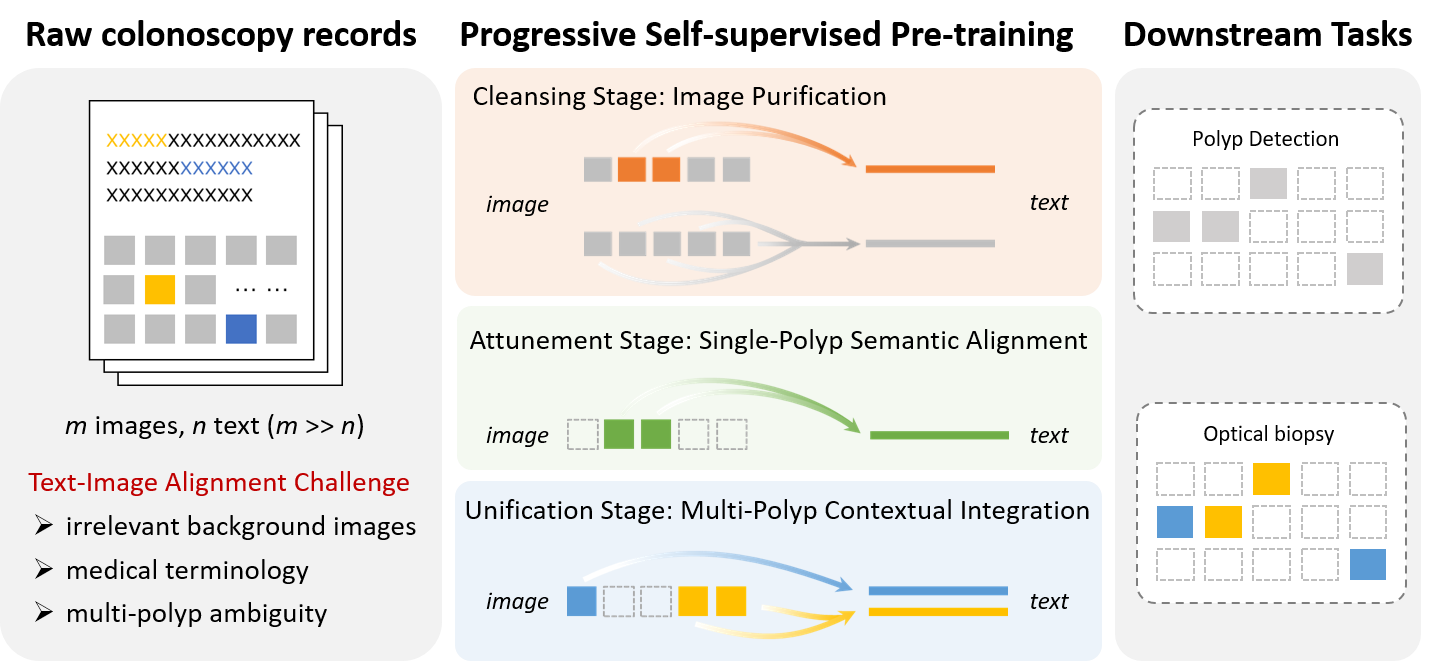}
\caption{Challenges of pre-training on colonoscopy records and the Endo-CLIP solution.} \label{fig:architecture1}
\end{figure}

Although initial efforts have been made \cite{wang2023foundation,batic2024endovit}, adapting CLIP to colonoscopy records presents unique and substantial challenges. The inherent complexities of colonoscopy data create a fundamental mismatch between images and text~\cite{wang2023knowledge}. In routine clinical practice, a single diagnostic report corresponds to numerous endoscopic images (often around 100 per examination), yet only a small fraction depict clinically significant findings (polyps), while the vast majority show normal mucosa.  This creates a critical "needle-in-a-haystack" problem: accurately associating the few relevant polyp images with the corresponding, often brief, textual descriptions within a lengthy report.  Furthermore, the considerable morphological variability of polyps – in size, shape, texture, and color – demands that the model effectively interpret nuanced medical terminology. Finally, the frequent occurrence of multiple polyps within a single examination (approximately 60\% in our cohort) introduces overlapping visual cues and potentially ambiguous textual descriptions, complicating the establishment of precise one-to-one correspondences between individual polyp images and segments of the report. To address these unique challenges, we present Endo-CLIP, a novel self-supervised framework for image-text alignment in raw colonoscopy records (Fig.~\ref{fig:architecture1}). The pre-trained Endo-CLIP models serves as backbones for downstream tasks, achieving high accuracy and robustness in polyp detection and classification, even under zero-shot and few-shot conditions.

\section{Method}
\label{sec:method}

\subsection{Model Overview}
\begin{figure}[h!]
\centering
\includegraphics[scale=0.5]{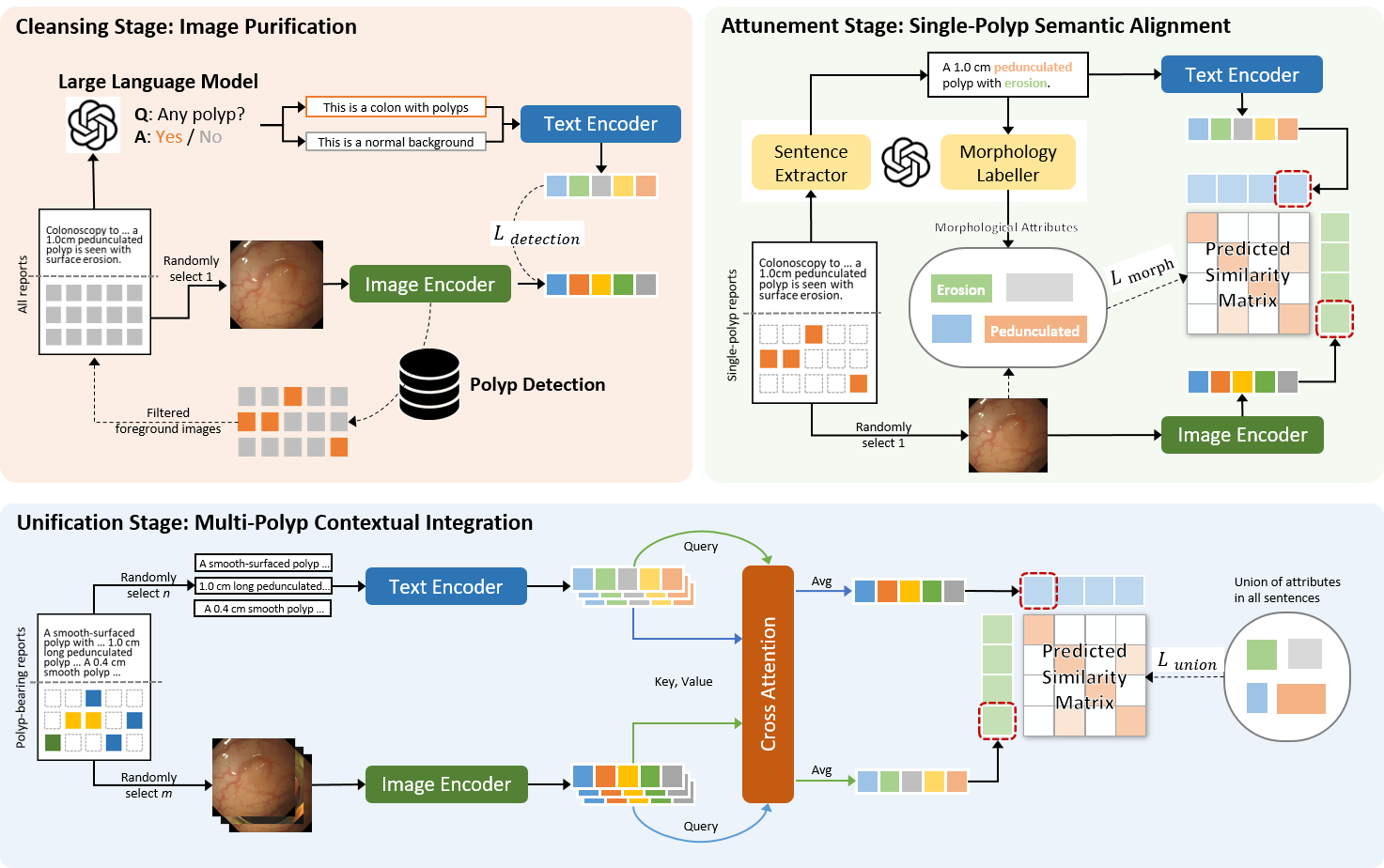}
\caption{Overview of Endo-CLIP: a progressive self-supervised pre-training framework that refines endoscopic image–text alignment.}
\label{fig:architecture2}
\end{figure}

Endo-CLIP (Fig.~\ref{fig:architecture2}) is a progressive self-supervised learning framework for endoscopic image-text alignment, addressing challenges like absent image-text pairs, background dominance, and variable polyp descriptions in colonoscopy data.  It comprises three stages: (1) Cleansing: Filters non-informative background frames using diagnostic sentences indicating polyp presence/absence. (2) Attunement: Enhances fine-grained polyp representation by enforcing semantic consistency between single-polyp images and their morphological attributes. (3) Unification: Resolves multi-polyp matching ambiguity via a cross-attention mechanism, dynamically associating polyp images with corresponding text descriptions and integrating contextual relationships.


\vspace{-1em}

\subsection{Cleansing Stage: Image Purification}
In the first stage, we identify polyp-bearing images from raw colonoscopy records, mitigating the noise introduced by normal bowel images and enhancing the quality of subsequent learning. We employ LLMs to preprocess each patient’s report \(R_i\), removing non-diagnostic content and extracting key descriptions for each polyp. Given \(N\) medical cases \(\{X_i, R_i\}\), where \(X_i \in \mathbb{R}^{K_i \times H \times W \times C}\) contains \(K_i\) images and \(R_i\) consists of a variable number of diagnostic sentences, we randomly sample one image \(I_i\) and encode it using a vision encoder \(\mathrm{E}_v\) as $ \mathbf{v}_i = \mathrm{E}_v(I_i)$.
If \(R_i\) describes polyps, we use the standardized sentence \texttt{`This is a colon with polyps.'}; otherwise, \texttt{`This is a normal background.'}. The selected sentence \(S_i\) is then encoded using a text encoder \(\mathrm{E}_t\) as $\mathbf{t}_i = \mathrm{E}_t(S_i)$.

To associate patient-level images with their corresponding textual descriptions, we employ a CLIP-style contrastive loss. We gather all \(\mathbf{v}_i\) and \(\mathbf{t}_i\) for a batch of \(N\) patients into matrices \(\mathbf{V}, \mathbf{T} \in \mathbb{R}^{N \times d}\). Let \(\mathrm{cosim}(\cdot,\cdot)\) denote the cosine similarity function applied row-wise, yielding two \(N \times N\) matrices:
\begin{equation}
\label{eq:cosim}
\mathbf{S}^{(v \to t)} = \mathrm{cosim}(\mathbf{V}, \mathbf{T}) \in \mathbb{R}^{N \times N}, 
\quad
\mathbf{S}^{(t \to v)} = \mathrm{cosim}(\mathbf{T}, \mathbf{V}) \in \mathbb{R}^{N \times N}.
\end{equation}
Following the InfoNCE formulation \cite{oord2018representation}, the detection loss is
\begin{equation}
\label{eq:contrast_loss_1}
\mathcal{L}_{\mathrm{detection}} = \frac{1}{2} 
\Bigl[ \mathrm{CE}\bigl(\mathbf{S}^{(v \to t)}, \mathbf{Y}^{(v \to t)}\bigr) 
     + \mathrm{CE}\bigl(\mathbf{S}^{(t \to v)}, \mathbf{Y}^{(t \to v)}\bigr) \Bigr],
\end{equation}
where \(\mathrm{CE}(\cdot, \cdot)\) is the cross-entropy loss computed on softmax-normalized similarities, and \(\mathbf{Y}^{(v \to t)}, \mathbf{Y}^{(t \to v)}\) are one-hot labels indicating positive matches along the diagonal. Since each image \(I_i\) is randomly sampled, many polyp-positive cases may contain normal bowel regions, introducing noise into the initial training.

To refine detection, we filter polyp-bearing images using a prevalence-guided selection strategy \cite{fang2023data}. Specifically, for polyp-positive patients, we compute the cosine similarity between each image and standardized polyp-positive and polyp-negative textual representations, followed by a softmax operation to obtain the probability of each image containing a polyp. 
Based on the known polyp prevalence, we retain the corresponding top percentage of highest-probability frames as polyp-bearing images. 
For polyp-negative patients, a randomly sampled frame remains as the negative example. We then reapply the loss in Eq.~\eqref{eq:contrast_loss_1} using these filtered images, initializing from the first-round model. This two-round contrastive learning approach enhances the \(\mathrm{E}_v\) as a polyp detector. The final outcome is a clean dataset of polyp-bearing images that sets the stage for fine-grained polyp representation learning in subsequent stages.

\subsection{Attunement Stage: Single-Polyp Semantic Alignment}
In the second stage, we focus on single-polyp cases to enhance fine-grained polyp representation learning by incorporating clinically defined morphological features. This approach mitigates false negatives arising from rigid one-hot supervision, which may overlook semantically similar pairs. We employ an LLM to extract morphological attributes of polyps from each diagnostic sentence \(S_i\), following clinical guidelines. 
For the \(i\)-th single-polyp case, these attributes are encoded into two multi-hot vectors \(\mathbf{a}_v^i\) and \(\mathbf{a}_t^i \), ensuring semantic consistency between the polyp image \(I_i\) and its textual description \(S_i\). Collecting these vectors for a batch of \(N\) single-polyp cases, we form two matrices \(\mathbf{A}_v \) and \(\mathbf{A}_t\), where row \(i\) corresponds to \((\mathbf{a}_v^i)^\top\) or \((\mathbf{a}_t^i)^\top\).
From these attribute matrices, we compute the morphological similarity matrices \(\mathbf{M}^{(v \to t)}\) and \(\mathbf{M}^{(t \to v)}\), both in \(\mathbb{R}^{N \times N}\). Each entry \(\mathbf{M}^{(v \to t)}_{ij}\) is computed as the cosine similarity between \(\mathbf{a}_v^i\) and \(\mathbf{a}_t^j\), representing the morphological alignment between the \(i\)-th image attribute vector and the \(j\)-th text attribute vector.

Meanwhile, the model produces similarity matrices \(\mathbf{S}^{(v \to t)}\) and \(\mathbf{S}^{(t \to v)}\) for the same batch of \(N\) single-polyp cases, using the image–text embeddings described in Eq.~\eqref{eq:cosim}. To integrate morphological knowledge, we replace the one-hot labels from Eq.~\eqref{eq:contrast_loss_1} with the soft targets \(\mathbf{M}^{(v \to t)}\) and \(\mathbf{M}^{(t \to v)}\). Formally,
\begin{equation}
\label{eq:phase2_loss}
\mathcal{L}_{\mathrm{morph}} = \frac{1}{2}
\Bigl[
\mathrm{CE}\bigl(\mathbf{S}^{(v \to t)}, \mathbf{M}^{(v \to t)}\bigr)
\;+\;
\mathrm{CE}\bigl(\mathbf{S}^{(t \to v)}, \mathbf{M}^{(t \to v)}\bigr)
\Bigr].
\end{equation}
By incorporating morphological attributes into the contrastive objective, we reduce false negatives and achieve finer medical semantic alignment.

\subsection{Unification Stage: Multi-Polyp Contextual Integration}
In the third stage, we refine representation learning for multi-polyp cases, where lesion-text associations are ambiguous. We introduce a polyp-level cross-attention mechanism \cite{vaswani2017attention,wang2022multi}, allowing each image and text embedding to attend to all possible matches. The attended features are then aggregated at the patient level, enabling holistic alignment through semantic similarity learning.

Let \(\{\mathbf{v}_{i}^1,\ldots,\mathbf{v}_{i}^{K_i}\}\) be the polyp image embeddings for patient \(i\), where \(K_i\) is the number of images, and let \(\{\mathbf{t}_{i}^1,\ldots,\mathbf{t}_{i}^{L_i}\}\) be the corresponding textual embeddings, where \(L_i\) is the number of descriptive sentences. Each polyp embedding \(\mathbf{v}_{i}^k\) attends to all textual embeddings \(\{\mathbf{t}_{i}^\ell\}\), and vice versa, using
\begin{equation}
\label{eq:phase3_cross}
\alpha_{i}^{k,\ell} 
= \mathrm{softmax}\Bigl(
\tfrac{(\mathbf{W}_Q\,\mathbf{v}_{i}^k)^\top\,(\mathbf{W}_K\,\mathbf{t}_{i}^\ell)}{\sqrt{d}}
\Bigr),
\quad
\mathbf{o}_{i,k}^{(v \to t)} 
= \sum_{\ell=1}^{L_i}
\alpha_{i}^{k,\ell}\,\bigl(\mathbf{W}_V\,\mathbf{t}_{i}^\ell\bigr),
\end{equation}
where \(\mathbf{W}_Q,\mathbf{W}_K,\mathbf{W}_V\) are learnable projection matrices. Symmetrically, each textual embedding \(\mathbf{t}_{i}^\ell\) attends over \(\{\mathbf{v}_{i}^k\}\) to produce \(\mathbf{o}_{i,\ell}^{(t \to v)}\). To form patient-level representations, we aggregate polyp-level features by averaging the updated image and text embeddings and then stack these aggregated embeddings from all patients into matrices \(\mathbf{\hat{V}}\) and \(\mathbf{\hat{T}} \in \mathbb{R}^{N \times d}\), enabling batch-wise similarity computation.

To incorporate lesion-specific details, we unify the morphological attributes of all \(L_i\) polyps into a single multi-hot label vector, representing the full spectrum of lesion characteristics for each patient. This ensures that our patient-level contrastive learning is guided by comprehensive morphological supervision. We then construct the soft target matrix \(\mathbf{\hat{M}}\).

We compute row-wise cosine similarities \(\mathbf{\hat{S}}^{(v \to t)}\) and \(\mathbf{\hat{S}}^{(t \to v)}\) with \(\mathbf{\hat{V}}\) and \(\mathbf{\hat{T}}\) according to Eq.~\eqref{eq:cosim}, and define the Stage~3 loss as:
\begin{equation}
\label{eq:stage3_loss}
\mathcal{L}_{\mathrm{union}}
= \frac{1}{2}\Bigl[
\mathrm{CE}\bigl(\mathbf{\hat{S}}^{(v \to t)}, \mathbf{\hat{M}}^{(v \to t)}\bigr)
+
\mathrm{CE}\bigl(\mathbf{\hat{S}}^{(t \to v)}, \mathbf{\hat{M}}^{(t \to v)}\bigr)
\Bigr].
\end{equation}

By aggregating multi-polyp features at both the polyp and patient levels, we mitigate matching ambiguity in multi-lesion scenarios while maintaining the fine-grained semantic alignment from the Attunement Stage.

\section{Experiment}

\subsection{Implementation Details}
The vision encoder adopts the ViT-B/32 version of the Vision Transformer \cite{dosovitskiy2020image}, while the text encoder utilizes a base Transformer architecture, both initialized with pre-trained CLIP weights. The model is optimized using AdamW with a learning rate of 5e-5. Training is conducted on eight NVIDIA A800 GPUs, with batch sizes of 32 for stage one and stage two, and 8 for stage three. Each stage is trained for 100 epochs, with the first 50 epochs dedicated to warm-up. 

\noindent\textbf{Pre-training Dataset:}
Our dataset comprises~13k colorectal examination cases with~874k endoscopic images and corresponding diagnostic reports. Polyp-positive cases account for~5.8k, including~2.5k multi-polyp cases (43\% of positives). 

\subsection{Downstream Task Setup}
\noindent\textbf{Datasets and Evaluation Metrics:}
We contribute and publicly release EndoReport50, a dataset of 84 polyps from 50 patients with detailed lesion annotations, including textual descriptions, morphological attributes of polyps (nine clinically relevant aspects annotated by clinicians) and pathology reports, and conduct experiments on it to evaluate our model. We consider two classification tasks: 1) Polyp Detection, distinguishing normal bowel images from polyp-containing ones, using a highly imbalanced test set of 2,308 normal and 608 polyp images. 2) Polyp Malignancy Classification, where polyps are categorized as benign (Vienna = 1) or malignant (Vienna = 3,4,5). Each polyp is represented by its most clinically indicative image, yielding 16 benign and 68 malignant samples. 

For both tasks, we report the Area Under the Receiver Operating Characteristic Curve (AUROC) and Area Under the Precision-Recall Curve (AUPR) to evaluate model performance under class imbalance. We compare models in zero-shot and few-shot settings, with training data ratios of 5\% and 10\% for polyp detection, and 10\% and 20\% for malignancy classification. 

\noindent\textbf{Comparison Methods:}
We compare our method with baselines covering supervised, vision contrast, and vision-language contrastive learning. \textbf{Random} is a ResNet-50 \cite{he2016deep} trained from scratch, while \textbf{ImageNet} is pretrained on ImageNet. \textbf{DINO} \cite{caron2021emerging} and \textbf{SSCD} \cite{pizzi2022self} use self-supervised learning, with SSCD tailored for medical imaging, and both of them pretrained on our dataset. \textbf{CLIP} \cite{radford2021learning} learns from large-scale image-text pairs for zero-shot transfer. All models are evaluated under the same setup for fair comparison.

\subsection{Experimental Results}

\noindent\textbf{Comparison with baseline methods.}
Table~\ref{tab:downstream_comparison} compares the performance of different methods on polyp detection and malignancy classification. Our Endo-CLIP models consistently outperform the baselines across all evaluation settings. For Polyp Detection, Endo-CLIP surpassing CLIP by +9.93\% AUROC in zero-shot and maintaining the lead in few-shot scenarios. This highlights the effectiveness of our two-round contrastive filtering in Stage 1. Self-supervised models like DINO and SSCD lag behind, underscoring the benefits of multimodal pretraining. For Malignancy Classification, Endo-CLIP attains state-of-the-art performance with 92.13\% AUROC in zero-shot, outperforming CLIP by +13.91\%. Its advantage persists in 10\% and 20\% few-shot settings, confirming the impact of morphology-aware learning and cross-attention in Stages 2 and 3.

\begin{table}[h]
\centering
\caption{Performance comparison of different methods on polyp detection and malignancy classification tasks. "-" indicates that the model does not support zero-shot learning. 
Endo-CLIP* refers to the model variant where the Cleansing Stage is used for polyp detection, while the Unification Stage is applied for malignancy classification.}
\label{tab:downstream_comparison}
\resizebox{\columnwidth}{!}{%
\begin{tabular}{l|cccccc|cccccc}
\hline
\multirow{3}{*}{Models}     & \multicolumn{6}{c|}{Polyp Detection}                                                                                                                                                                         & \multicolumn{6}{c}{Malignancy Classification}                                                                                                                                                                \\ \cline{2-13} 
                            & \multicolumn{3}{c|}{AUROC}                                                                                      & \multicolumn{3}{c|}{AUPR}                                                                  & \multicolumn{3}{c|}{AUROC}                                                                                      & \multicolumn{3}{c}{AUPR}                                                                   \\ \cline{2-13} 
                            & \multicolumn{1}{c|}{Zero-shot}      & \multicolumn{1}{c|}{5\%}            & \multicolumn{1}{c|}{10\%}           & \multicolumn{1}{c|}{Zero-shot}      & \multicolumn{1}{c|}{5\%}            & 10\%           & \multicolumn{1}{c|}{Zero-shot}      & \multicolumn{1}{c|}{10\%}           & \multicolumn{1}{c|}{20\%}           & \multicolumn{1}{c|}{Zero-shot}      & \multicolumn{1}{c|}{10\%}           & 20\%           \\ \hline
Random                      & \multicolumn{1}{c|}{-}              & \multicolumn{1}{c|}{55.61}          & \multicolumn{1}{c|}{59.34}          & \multicolumn{1}{c|}{-}              & \multicolumn{1}{c|}{24.75}          & 26.57          & \multicolumn{1}{c|}{-}              & \multicolumn{1}{c|}{48.02}          & \multicolumn{1}{c|}{53.16}          & \multicolumn{1}{c|}{-}              & \multicolumn{1}{c|}{79.80}          & 83.29          \\
ImageNet                    & \multicolumn{1}{c|}{-}              & \multicolumn{1}{c|}{73.45}          & \multicolumn{1}{c|}{78.31}          & \multicolumn{1}{c|}{-}              & \multicolumn{1}{c|}{42.95}          & 50.05          & \multicolumn{1}{c|}{-}              & \multicolumn{1}{c|}{48.17}          & \multicolumn{1}{c|}{54.40}          & \multicolumn{1}{c|}{-}              & \multicolumn{1}{c|}{82.06}          & 86.79          \\
Dino                        & \multicolumn{1}{c|}{-}              & \multicolumn{1}{c|}{67.11}          & \multicolumn{1}{c|}{68.96}          & \multicolumn{1}{c|}{-}              & \multicolumn{1}{c|}{35.97}          & 38.36          & \multicolumn{1}{c|}{-}              & \multicolumn{1}{c|}{51.58}          & \multicolumn{1}{c|}{57.55}          & \multicolumn{1}{c|}{-}              & \multicolumn{1}{c|}{82.98}          & 85.93          \\
SSCD                        & \multicolumn{1}{c|}{-}              & \multicolumn{1}{c|}{57.07}          & \multicolumn{1}{c|}{62.85}          & \multicolumn{1}{c|}{-}              & \multicolumn{1}{c|}{27.15}          & 31.91          & \multicolumn{1}{c|}{-}              & \multicolumn{1}{c|}{52.12}          & \multicolumn{1}{c|}{53.41}          & \multicolumn{1}{c|}{-}              & \multicolumn{1}{c|}{83.51}          & 84.96          \\
CLIP                        & \multicolumn{1}{c|}{66.45}          & \multicolumn{1}{c|}{70.25}          & \multicolumn{1}{c|}{75.00}          & \multicolumn{1}{c|}{31.27}          & \multicolumn{1}{c|}{37.80}          & 45.41          & \multicolumn{1}{c|}{45.22}          & \multicolumn{1}{c|}{45.63}          & \multicolumn{1}{c|}{47.39}          & \multicolumn{1}{c|}{78.18}          & \multicolumn{1}{c|}{80.53}          & 81.28          \\ \hline
Endo-CLIP\textsuperscript{*} & \multicolumn{1}{c|}{\textbf{76.38}} & \multicolumn{1}{c|}{\textbf{79.08}} & \multicolumn{1}{c|}{\textbf{82.38}} & \multicolumn{1}{c|}{\textbf{45.60}} & \multicolumn{1}{c|}{\textbf{50.38}} & \textbf{54.37} & \multicolumn{1}{c|}{\textbf{72.79}} & \multicolumn{1}{c|}{\textbf{58.72}} & \multicolumn{1}{c|}{\textbf{62.73}} & \multicolumn{1}{c|}{\textbf{92.13}} & \multicolumn{1}{c|}{\textbf{87.68}} & \textbf{89.61} \\ \hline
\end{tabular}%
}
\end{table}

\begin{figure}[h!]
\centering
\includegraphics[width=\textwidth]{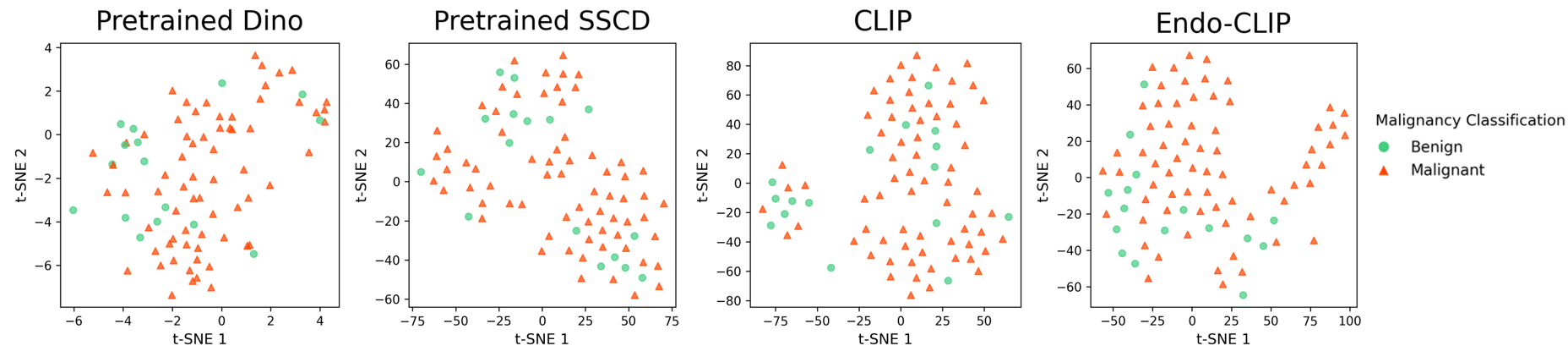}
\caption{t-SNE \cite{van2008visualizing} visualization of feature distributions for malignancy classification. Each point represents a polyp image in the feature space, with green circles for benign cases and red triangles for malignant cases.} \label{fig:tsne}
\end{figure}

\noindent\textbf{Feature Visualization.}
To gain deeper insights into the feature representations learned by different models, we utilize t-SNE \cite{van2008visualizing} visualization to project high-dimensional feature embeddings into a two-dimensional space. Figure \ref{fig:tsne} illustrates the t-SNE projections of feature distributions for malignancy classification, comparing DINO, SSCD, CLIP, and Endo-CLIP.
The DINO and SSCD models produce dispersed and overlapping clusters, suggesting their limited ability to distinguish between benign and malignant cases. CLIP, which benefits from large-scale image-text pre-training, demonstrates improved separation, although some overlap between classes remains. In contrast, Endo-CLIP achieves the most distinct clustering, with well-defined boundaries between benign and malignant polyps.
These results emphasize the effectiveness of hierarchical contrastive learning and morphology-aware supervision in refining polyp representations. 

\noindent\textbf{Ablation Study.}
Table~\ref{tab:ablation} examines the impact of dataset composition and key methodological components on malignancy classification. Data ablation shows that incorporating both single-polyp (SP) and multi-polyp (MP) cases leads to improved performance, indicating that a diverse dataset aids generalization. On the methodological side, replacing Morphology Contrastive Learning (MC) with InfoNCE loss \cite{oord2018representation} results in degraded performance, suggesting that morphology-aware supervision provides useful semantic guidance. Similarly, replacing the Cross-Attention (CA) mechanism with simple averaging weakens multi-polyp feature integration, underscoring the role of explicit attention modeling. The full model achieves the highest scores, validating the effectiveness of our hierarchical framework in refining polyp representations.

\begin{table}[]
\centering
\caption{Results of ablation studies of Endo-CLIP.}
\label{tab:ablation}
\resizebox{\columnwidth}{!}{%
\begin{tabular}{l|cc|cc|cccccc}
\hline
\multirow{3}{*}{Model}    & \multicolumn{2}{c|}{Data}                 & \multicolumn{2}{c|}{Method}               & \multicolumn{6}{c}{Malignancy Classification}                                                                                                                                                                           \\ \cline{2-11} 
                          & \multirow{2}{*}{SP} & \multirow{2}{*}{MP} & \multirow{2}{*}{MC} & \multirow{2}{*}{CA} & \multicolumn{3}{c|}{AUROC}                                                                                      & \multicolumn{3}{c}{AUPR}                                                                              \\
                          &                     &                     &                     &                     & Zero-shot                           & 10\%                                & \multicolumn{1}{c|}{20\%}           & Zero-shot                           & 10\%                                & 20\%                      \\ \hline
\multirow{5}{*}{Endo-CLIP} &                     &                     &                     &                     & \multicolumn{1}{c|}{48.81}          & \multicolumn{1}{l|}{38.47}          & \multicolumn{1}{l|}{48.28}          & \multicolumn{1}{c|}{80.17}          & \multicolumn{1}{l|}{77.21}          & \multicolumn{1}{l}{81.09} \\
                          & \checkmark          &                     &                     &                     & \multicolumn{1}{c|}{50.00}          & \multicolumn{1}{c|}{50.13}          & \multicolumn{1}{c|}{52.35}          & \multicolumn{1}{c|}{81.83}          & \multicolumn{1}{c|}{81.62}          & 83.25                     \\
                          & \checkmark          &                     & \checkmark          &                     & \multicolumn{1}{c|}{39.71}          & \multicolumn{1}{c|}{54.23}          & \multicolumn{1}{c|}{59.36}          & \multicolumn{1}{c|}{78.25}          & \multicolumn{1}{c|}{84.78}          & 87.32                     \\
                          & \checkmark          & \checkmark          & \checkmark          &                     & \multicolumn{1}{c|}{48.35}          & \multicolumn{1}{c|}{58.05}          & \multicolumn{1}{c|}{55.17}          & \multicolumn{1}{c|}{84.33}          & \multicolumn{1}{c|}{84.76}          & 86.50                     \\
                          & \checkmark          & \checkmark          & \checkmark          & \checkmark          & \multicolumn{1}{c|}{\textbf{72.79}} & \multicolumn{1}{c|}{\textbf{58.72}} & \multicolumn{1}{c|}{\textbf{62.73}} & \multicolumn{1}{c|}{\textbf{92.13}} & \multicolumn{1}{c|}{\textbf{87.68}} & \textbf{89.61}            \\ \hline
\end{tabular}%
}
\end{table}

\section{Conclusion}
We introduced Endo-CLIP, a progressive self-supervised learning framework tailored for pre-training on raw colonoscopy records. To address the challenges of image-text alignment, we propose a three-stage approach: (1) Image purification filters out non-informative frames, (2) single-polyp semantic aligment leverages clinically defined attributes for fine-grained semantic learning, and (3) multi-polyp contextual integration applies cross-attention to resolve complex image-text associations at the patient level. Experimental results demonstrate that Endo-CLIP outperforms baselines in both zero-shot and few-shot settings for polyp detection and malignancy classification. By enhancing the alignment of endoscopic images and textual reports, Endo-CLIP lays a foundation for more accurate and clinically meaningful endoscopic analysis.

\bibliographystyle{splncs04}
\bibliography{ref}

\end{document}